\documentclass[runningheads]{llncs}

\usepackage{eccv}

\usepackage{eccvabbrv}

\usepackage{graphicx}
\usepackage{booktabs}

\usepackage[accsupp]{axessibility}  

\newcommand*\samethanks[1][\value{footnote}]{\footnotemark[#1]}

\usepackage{hyperref}

\usepackage{orcidlink}

\usepackage{colortbl}

\begin{document}

\title{Progressive Classifier and Feature Extractor Adaptation for Unsupervised Domain Adaptation on Point Clouds} 

\titlerunning{PCFEA}

\author{Zicheng Wang\inst{1, 2} \and Zhen Zhao\inst{2} \and Yiming Wu\inst{1} \and Luping Zhou\inst{2}\thanks{Corresponding authors.} \and Dong Xu\inst{1}\samethanks \\
}

\authorrunning{Z. Wang et al.}

\institute{The University of Hong Kong \and The University of Sydney}

\maketitle

\begin{abstract}
    Unsupervised domain adaptation (UDA) is a critical challenge in the field of point cloud analysis.
    Previous works tackle the problem either by feature extractor adaptation to enable a shared classifier to distinguish domain-invariant features, or by classifier adaptation to evolve the classifier to recognize target-styled source features to increase its adaptation ability.
    However, by learning domain-invariant features, feature extractor adaptation methods fail to encode semantically meaningful target-specific information, while classifier adaptation methods rely heavily on the accurate estimation of the target distribution.
    In this work, we propose a novel framework that deeply couples the classifier and feature extractor adaption for 3D UDA, dubbed Progressive Classifier and Feature Extractor Adaptation (PCFEA). Our PCFEA conducts 3D UDA from two distinct perspectives: macro and micro levels. On the macro level, we propose a progressive target-styled feature augmentation (PTFA) that establishes a series of intermediate domains to enable the model to progressively adapt to the target domain. Throughout this process, the source classifier is evolved to recognize target-styled source features (\ie, classifier adaptation). On the micro level, we develop an intermediate domain feature extractor adaptation (IDFA) that performs a compact feature alignment to encourage the target-styled feature extraction gradually. 
    In this way, PTFA and IDFA can mutually benefit each other: IDFA contributes to the distribution estimation of PTFA while PTFA constructs smoother intermediate domains to encourage an accurate feature alignment of IDFA.
    We validate our method on popular benchmark datasets, where our method achieves new state-of-the-art performance.
    Our code is available at \url{https://github.com/xiaoyao3302/PCFEA}.
\end{abstract}




\section{Introduction}
\label{sec:intro}

Deep learning on point clouds has revolutionized various real-world applications, including autonomous driving and robotics~\cite{shi2019pointrcnn, shi2020pv, wu2023virtual, rukhovich2023tr3d, das2018embodied, fang2020graspnet, liu2024lta}. However, a significant challenge arises from the domain shift, where a model trained on one set of data fails to perform well in a new scenario due to differences in scanning devices or geometry variations. To address this issue, numerous unsupervised domain adaptation methods have been developed to tackle the cross-domain point cloud recognition problem~\cite{zou2021geometry, zhang2018collaborative, qin2019pointdan, ghifary2016deep, li2013learning}. 

\begin{figure}[t]
\centering
\includegraphics[width=0.9\linewidth]{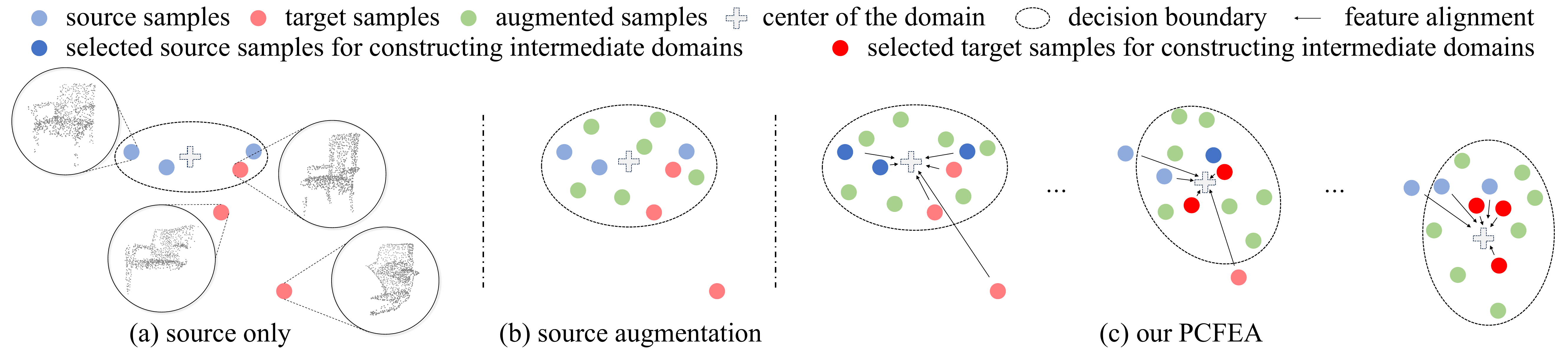}
\caption{Illustration of our proposed progressive classifier and feature extractor adaptation (PCFEA) approach using samples from the same category in different domains.
(a) The gaps between different target samples to the source domain are different, while a vanilla model trained solely with source data cannot recognize certain target samples with a large gap to the source domain. 
(b) A model trained with augmented source samples will improve the generalization of the model.
(c) Considering the uniqueness of each sample, we select part of the source and target samples, \ie, the dark-colored solid ones, to progressively generate new intermediate domains toward the target domain. We construct feature augmentations and perform feature alignment towards the new intermediate domain and the model trained will gradually approach the target domain.} 
\label{fig_demo}
\end{figure}

Most point cloud UDA methods focus on \textit{feature extraction adaptation}, which targets at extracting compact domain-invariant features that can be classified by a shared classifier~\cite{qin2019pointdan, wang2021cross, achituve2021self, zou2021geometry, liang2022point, shen2022domain, fan2022self, cardace2023self}. 
However, these methods fail to encode semantic meaningful target-specific information and thus are far from optimal for target sample recognition~\cite{wang2023domain}.
Currently, some 2D methods have been proposed to tackle the domain adaptation problem from a new perspective of \textit{classifier adaptation}, which encourages the classifier to recognize target-styled source features to increase the adaptation ability of the classifier~\cite{li2021transferable}. However, these methods rely heavily on the accurate estimation of the target distribution, and incorrect target estimation will degrade the adaptation performance.
More importantly, due to the scanning noise, as shown in~\cref{fig_demo}~(a), in the context of point cloud adaptation, the distances between various target samples and the source domain exhibit significant disparities~\cite{chen2019progressive}. These disparities result in considerable variations in prediction accuracy in the target domain, leading to unreliable estimation of the target distribution in classifier adaptation methods. As a result, it is inappropriate to treat each target sample equally~\cite{li2021transferable}. 
This issue, although overlooked by previous studies, is crucial in 3D point-cloud UDA.

\begin{figure}[t]
\centering
\includegraphics[width=0.65\linewidth]{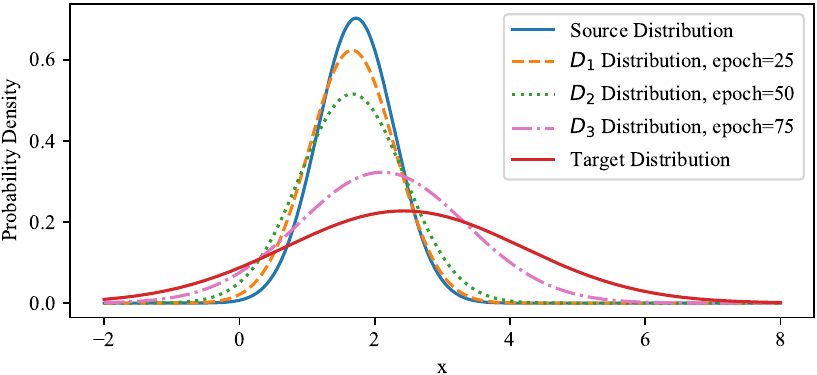}
\caption{Visualization of the changing process of the distributions of the source, the target, and different intermediate domains. We use samples from the same category under the adaptation scenario ModelNet-10 $\rightarrow$ ScanNet-10 as an example. $D_k$ indicates the constructed intermediate domain during the $k$-th stage.}
\label{fig_distribution}
\end{figure}

Based on our observation, in this work, we propose a new progressive classifier and feature extractor adaptation (PCFEA) method to tackle the cross-domain point cloud recognition problem. 
In particular, \textit{being the first to consider the sample uniqueness in 3D UDA}, we first come up with a new progressive target-styled feature augmentation (PTFA) strategy for \textit{classifier adaptation}, which involves creating a series of intermediate domains between the source domain and the target domain, and we encourage the model to progressively approach the intermediate domains to avoid direct adaptation across a significant domain gap~\cite{hsu2020progressive}. 
During the process, we select samples based on their prediction scores, progressively decreasing the selected source samples and increasing the selected target samples. Then we generate intermediate domains gradually moving towards the target domain based on the selected samples, \ie, dark-colored solid circles in~\cref{fig_demo}~(c). 
By progressively performing the feature augmentations on the source features towards the new intermediate domain, we encourage the model to progressively recognize the augmented source features, which gradually improves the adaptation ability of the classifier. \cref{fig_distribution} clearly demonstrates that the distribution of the new intermediate domains can progressively approach the target distribution. 
In addition, considering that the prediction accuracy of the target samples will greatly influence the estimation of the distribution~\cite{lifshitz2021sample}, based on our progressively estimated intermediate domains, we further introduce a new intermediate domain feature extractor adaptation (IDFA) strategy. We encourage features extracted from the source samples and the target samples to approach the mean of the distribution of the intermediate domain, which encourages a compact feature alignment, boosting the adaptation performance.

Notably, our PTFA and IDFA are deeply integrated, mutually benefiting each other. 
On one hand, IDFA enforces a compact feature alignment and enhances the distribution estimation in TSFA. 
On the other hand, PTFA enables the model to gradually reduce the domain gap and improve the adaptation ability of the classifier, thus reducing the wrong alignment risk of IDFA caused by incorrect predictions. 
In summary, our contributions are threefold:
\begin{itemize}
\item We propose a new progressive classifier and feature extractor adaptation (PCFEA) approach that deeply couples the classifier and feature extractor adaption to tackle unsupervised domain adaptation on point clouds.
\item We propose a progressive target-styled feature augmentation (PTFA) strategy and an intermediate domain feature extractor adaptation (IDFA) strategy, which can mutually benefit each other.
\item Our method achieves state-of-the-art (SOTA) performance on 3D UDA benchmark datasets, surpassing the existing methods by a significant margin.
\end{itemize}

\section{Related Work}
\label{sec:related_work}

\subsection{Deep Learning on Point Clouds}
Deep learning has achieved great success~\cite{zhao2022dc, zhao2022lassl, zhao2023augmentation, zhao2023instance, wang2023conflict, zhao2021mgsvf, zhao2021memory, zhao2021video, zhao2022rbc, chu2023towards, chen2023neural, chen2022exploiting, chen2022lsvc, chen2021improving, zhao2023alternate}. Recently, inspired by PointNet~\cite{qi2017pointnet}, extensive deep models have been proposed~\cite{qi2017pointnet++, wang2019dynamic, wang2024pointramba} to use multi-layer perceptions (MLPs) for point cloud recognition. 
While these methods have achieved remarkable performance in different scenarios, a learned model using a certain set of training data can hardly perform well in a new scenario, which is known as the unsupervised domain adaptation (UDA) problem~\cite{zou2021geometry, shen2022domain, wang2023domain, wu2021mgh}.

\subsection{Feature Extraction Adaptation}
Various methods have been proposed to tackle the UDA problem on point clouds and most of these methods focus on the feature extractor adaptation. These methods can be categorized into two main categories, \ie, the adversarial-training (AT)-based methods~\cite{qin2019pointdan, wang2021cross} and self-supervised learning (SSL)-based methods~\cite{achituve2021self, zou2021geometry, liang2022point, shen2022domain, fan2022self, cardace2023self}. 

The AT-based methods aim at learning a feature extractor to extract domain-invariant features from both source samples and target samples that cannot be distinguished by a domain discriminator. However, adversarial training encounters stability issues and lacks a guarantee of semantically meaningful feature extraction~\cite{shen2022domain}.
The SSL-based methods aim at designing self-supervised tasks to enable the feature extractor to extract semantically meaningful features from the target samples, which are then delivered to the classifier supervised by source labels for classification. However, self-supervised tasks primarily facilitate the learning of low-level geometry features, which may be inadequate for high-level recognition tasks, and there is no theoretical proof ensuring that the extracted features from target samples using self-supervised tasks can be distinguished by the source-supervised classifier~\cite{rao2020global, rao2022pointglr, sauder2019self}.


Moreover, both the AT-based methods and SSL-based methods fail to encode semantic meaningful target-specific information, which can hardly be optimal for target sample recognition~\cite{wang2023domain}.

\subsection{Classification Adaptation}
Classifier adaptation is another typical adaptation method, which targets at increasing the adaptation ability of the classifier using feature augmentations.
ISDA~\cite{wang2021regularizing} is a representative 2D work that proposed to construct semantically meaningful feature augmentations with different directions to expand the decision region of the classifier so as to improve the generalization of the model. 
However, ISDA cannot handle large domain gaps as it can only construct feature augmentations based on the estimation of the distribution of the given training data~\cite{li2021transferable}. 
TSA~\cite{li2021transferable} is another representative 2D work that tackles the problem by using the estimated distribution of the target domain as the direction of the feature augmentations to construct target-styled feature augmentations and then perform classifier adaptation.
However, TSA directly takes all of the target data for distribution estimation, while in the 3D domain, the distance between different samples and the source domain varies significantly. Directly estimating the target distribution with all of the target predictions will inevitably lead to estimation bias, which will adversely affect the overall adaptation performance.

\subsection{Gradual domain alignment}
Various gradual domain alignment methods have been proposed to tackle the 2D UDA problem~\cite{kumar2020understanding, abnar2021gradual, zhang2021gradual, tang2021vicinal, yue2021prototypical}. However, most of these methods rely heavily on the predicted pseudo-labels and are vulnerable to incorrect predictions. In addition, the diverse scanning noise in point clouds makes 3D UDA more challenging, requiring more robust estimation and dedicated gradual domain alignment.

\begin{figure}[t]
\centering
\includegraphics[width=0.85\linewidth]{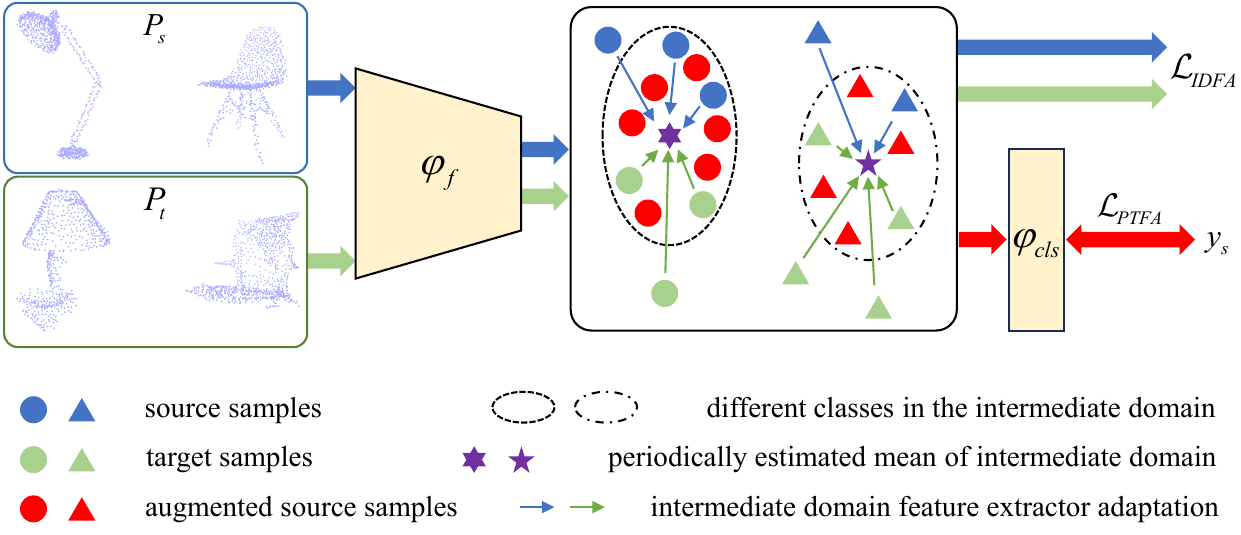}
\caption{An overview of our proposed method for unsupervised domain adaptation on point clouds. We progressively generate intermediate domains that gradually resemble the target domain. Taking one intermediate domain as an example, we first propose a progressive target-styled feature augmentation (PTFA) approach to construct semantically meaningful feature augmentations on the source features extracted by $\varphi_f$ towards the intermediate domain, which is then fed into the classifier $\varphi_{cls}$ to generate source predictions. The source predictions are supervised by the ground-truth labels $y_s$. Directly optimizing the corresponding $\mathcal{L}_{PTFA}$ enables the classifier to be adapted toward the intermediate domain. We further propose an intermediate domain feature extractor adaptation (IDFA) strategy to encourage the features extracted from the source samples and the target samples to approach the estimated mean of the intermediate domain using $\mathcal{L}_{IDFA}$. 
}
\label{fig_pipeline}
\end{figure}

\section{Method}
\label{sec:method}
In this section, we present our newly proposed progressive classifier and feature extractor adaptation (PCFEA) approach in detail, which consists of a progressive target-styled feature augmentation (PTFA) approach and an intermediate domain feature extractor adaptation (IDFA). First, we introduce the problem formulation of this work in~\cref{sec:formulation}. Then, we elaborate our PTFA approach, which consists of a progressive target-approaching (PTA) in~\cref{sec:PTA}, and a target-styled feature augmentation (TSFA) in~\cref{sec:TSFA}. We further present our IDFA strategy in~\cref{sec:IDA}. Finally, the overall training loss is given in~\cref{sec:overall_method} and the pipeline of our method is illustrated in~\cref{fig_pipeline}.

\subsection{Problem Formulation}
\label{sec:formulation}
Following previous works~\cite{qin2019pointdan, achituve2021self, fan2022self, shen2022domain}, 
we are given a set of source data $\boldsymbol{\mathbb{S}}=\left\{\boldsymbol{\mathcal{P}}_{n}^{s}, y_{n}^{s}\right\}_{n=1}^{N_{s}}$, indicating $N_{s}$ point clouds $\boldsymbol{\mathcal{P}}_n^{s}$ in total with their corresponding category labels $y_n^{s}$. Each point cloud sample $\boldsymbol{\mathcal{P}}_n^{s} \in \mathbb{R}^{m \times 3}$ consists of $m$ unordered points with their corresponding 3D coordinates. Similarly, we are given another set of target data $\boldsymbol{\mathbb{T}}=\left\{\boldsymbol{\mathcal{P}}_{n}^{t}, y_{n}^{t}\right\}_{n=1}^{N_{t}}$, while the target labels $y_{n}^{t}$ are only available during the inference stage. All of the samples from the source and the target domain share the same label space $\mathcal{Y} =\{1,2, \cdots, C\}$. We aim at learning a model $\varphi = \varphi_{f} \circ \varphi_{cls}$ that can perform well on the target domain, where $\varphi_{f}$ indicates the feature extractor and $\varphi_{cls}$ indicates the classifier.


Given the source data $\boldsymbol{\mathcal{P}}_{n}^{s}$, it is easy to extract the corresponding features $\boldsymbol{f}_n^s$ from each sample by $\boldsymbol{f}_n^s = \varphi_{f}\left(\boldsymbol{\mathcal{P}}_{n}^{s}\right)$. Similarly, we can extract target features $\boldsymbol{f}_n^t = \varphi_{f}\left(\boldsymbol{\mathcal{P}}_{n}^{t}\right)$ from the target data $\boldsymbol{\mathcal{P}}_{n}^{t}$. In this work, we construct two \textit{feature pools} $\boldsymbol{\mathbb{F}}_s = \left\{\boldsymbol{f}_n^s\right\}_{n=1}^{N_{s}}$ to store $\boldsymbol{f}_n^s$ and $\boldsymbol{\mathbb{F}}_t = \left\{\boldsymbol{f}_n^t\right\}_{n=1}^{N_{t}}$ to store $\boldsymbol{f}_n^t$. Given an extracted feature $\boldsymbol{f}_n$ and a classifier $\varphi_{cls}$, we can obtain the corresponding prediction $\boldsymbol{\xi}_{n}=\varphi_{cls}\left(\boldsymbol{f}_n\right)$. Therefore, we can construct two \textit{confidence score pools} $\mathbb{Z}_s = \left\{\zeta_n^s\right\}_{n=1}^{N_{s}}$ and $\mathbb{Z}_t = \left\{\zeta_n^t\right\}_{n=1}^{N_{t}}$, where $\zeta_n = \max_c \boldsymbol{\xi}_n$, $c$ indicates the index of the category. Finally, we construct two \textit{label pools} $\mathbb{H}_s = \left\{\eta_n^s\right\}_{n=1}^{N_{s}}$ and $\mathbb{H}_t = \left\{\eta_n^t\right\}_{n=1}^{N_{t}}$, where $\eta_n^s = y_n^s$ as we have ground-truth labels available for each source sample, and $\eta_n^t = \arg\max_c \boldsymbol{\xi}_n^t$, indicating the corresponding pseudo-label of each target sample. According to the corresponding $\eta_n$, we can divide the pools into category-wise feature pools as $\boldsymbol{\mathbb{F}}_s^c$ and $\boldsymbol{\mathbb{F}}_t^c$, and confidence score pools as $\boldsymbol{\mathbb{Z}}_s^c$ and $\boldsymbol{\mathbb{Z}}_t^c$. It is easy to estimate the mean and the covariance of the source and target distributions, \ie, $\boldsymbol{\mu}_s^c$ and $\boldsymbol{\Sigma}_s^c$, $\boldsymbol{\mu}_t^c$ and $\boldsymbol{\Sigma}_t^c$, according to the source labels or the target pseudo-labels.

\subsection{Progressive Target-Approaching}
\label{sec:PTA}   
Recall that we aim to construct feature augmentations on the source samples toward the target domain to perform classifier adaptation. The first issue to determine is the \textit{direction of the feature augmentations}. Different from the previous works like TSA~\cite{li2021transferable} that directly estimate the augmentation direction by $\Delta \boldsymbol{\mu}_c = \boldsymbol{\mu}_t^c - \boldsymbol{\mu}_s^c$, we argue that directly adapting the model across a significant domain gap may hamper the adaptation performance, as the estimation of the target distribution would be swayed by the incorrect target predictions, especially during the initial training stages. Moreover, the distances between various target samples and the source domain exhibit significant disparities, some hard samples that lie far away from the source domain may be ambiguous for the classification boundaries.
To address the concerns, we propose a new progressive target-approaching strategy to divide the whole training process into several stages with each stage consisting of $\tau$ training epochs. Considering that the total training epoch is $T$, the whole training process can be divided into $\lceil T / \tau \rceil$ stages, where $\lceil \cdot \rceil$ is a round-up operation. We construct one new intermediate domain $D_k$ between the source and the target domain for each stage. With the intermediate domains progressively approaching the target domain, the model is encouraged to approach the intermediate domains from the source domain toward the target domain. During each stage, we aim to select samples that are closest to the mean of the current intermediate domain. This operation ensures the domain gap between two consecutive intermediate domains is small, preserving the prediction accuracy of the target samples. Consequently, this approach enhances the estimation of the new intermediate domain estimation and contributes significantly to the overall adaptation performance. Intuitively, the distance between a sample and the mean of the intermediate domain can be reflected by the prediction score $\zeta_n$, and a higher $\zeta_n$ usually indicates that the sample may probably lie within the decision region of the classifier, with a short distance to mean of the intermediate domain. Consequently, we only select the samples with the highest $\zeta_n$ from both domains to construct the new intermediate domain.

Moreover, considering the inherent long-tailed distribution of the dataset, we propose a new ratio-based category-wise sample selection strategy to select a fixed ratio of samples from each category, aiming to pick $\sigma^s$ samples from each category of the source domain and $\sigma^t$ samples from each category of the target domain. Here $\sigma^s$ and $\sigma^t$ represent the proportion of the number of samples we wish to select from each category. To implement this, we sort $\mathbb{Z}_s^c$ and $\mathbb{Z}_t^c$ and keep the former $\sigma^s$ source samples and the former $\sigma^t$ source samples from each category. In addition, as our objective is to create target-styled feature augmentation, we advocate for the progressive convergence of the intermediate domain towards the target domain. To achieve this, we progressively decrease the number of the selected source samples and increase the number of the selected target samples, \ie, $\sigma_{k+1}^s = \sigma_k^s - \Delta \sigma^s$ and $\sigma_{k+1}^t = \sigma_k^t + \Delta \sigma^t$, where $k$ indicates the $k$-th intermediate domain construction stage and $\Delta \sigma$ indicates the proportional change during each stage. This approach results in a gradual shift by considering fewer source samples and more target samples to construct the intermediate domain, ensuring a progressive approach toward the target domain. It is easy to estimate the mean and covariance of the newly constructed intermediate domain as $\boldsymbol{\mu}_k^c$ and $\boldsymbol{\Sigma}_k^c$. During each training stage, the augmentation direction is estimated by $\Delta \boldsymbol{\mu}_k^c = \boldsymbol{\mu}_k^c - \boldsymbol{\mu}_s^c$.

\subsection{Target-Styled Feature Augmentation}
\label{sec:TSFA}
After determining the direction of the feature augmentation, the next issue to determine is the \textit{formulation of the feature augmentation}. 
Given an estimated distribution of the intermediate domain $N\left(\boldsymbol{\mu}_k^c, \boldsymbol{\Sigma}_k^c\right)$, it can be represented by $N\left(\boldsymbol{\mu}_s^c + \Delta \boldsymbol{\mu}_k^c, \boldsymbol{\Sigma}_k^c\right)$, where $\Delta \boldsymbol{\mu}_k^c = \boldsymbol{\mu}_k^c - \boldsymbol{\mu}_s^c$. Recall that the mean of a distribution stands for its semantics and the covariance of a distribution stands for its semantic variations. As we aim at constructing target-styled source features, we can generate $M$ augmentations sampled from a distribution $N\left(\Delta \boldsymbol{\mu}_t^c, \boldsymbol{\Sigma}_t^c\right)$ on a certain source feature $\boldsymbol{f}_n^s$. That is, augmented source features $ \widetilde{\boldsymbol{f}}_{n, m}^s, m=1 \cdots M$ confirm to a distribution of $N\left(\boldsymbol{f}_n^s + \Delta \boldsymbol{\mu}_k^c, \boldsymbol{\Sigma}_k^c\right)$, which share the same semantics and variations as the constructed intermediate domain. Following \cite{wang2021regularizing} and \cite{li2021transferable}, we also use a parameter $\lambda$ to control the strength of the augmentation, \ie, $\lambda \boldsymbol{\Sigma}_k^c$, to enable the semantic variants to approach the target distribution progressively, where $\lambda = \left(t/T\right)\times \lambda_0$, $t$ and $T$ indicate the current training epoch and the total training epoch, respectively. Considering that a linear classifier $\varphi_{cls}$ is defined by a weight matrix $\boldsymbol{W} = \left[\boldsymbol{w}_1, \boldsymbol{w}_2, \cdots, \boldsymbol{w}_C \right]^T$ and a bias vector $\boldsymbol{b} = \left[ b_1, b_2, \cdots, b_C\right]^T$, we can perform supervision on the augmented source features by minimizing $\mathcal{L}_M$ as:
\begin{equation}
    \mathcal{L}_M=\frac{1}{N_s} \sum_{n=1}^{N_s} \frac{1}{M} \sum_{m=1}^M-\log \left(\frac{e^{\boldsymbol{w}_{y_{n}^{s}}^{\top}  \widetilde{\boldsymbol{f}}_{n, m}^s+b_{y_{i}^{s}}}}{\sum_{c=1}^C e^{\boldsymbol{w}_c^{\top}  \widetilde{\boldsymbol{f}}_{n, m}^s+b_c}}\right).
\end{equation}
Directly constructing $M$ augmentations and optimizing $\mathcal{L}_M$ is difficult to implement. However, if we construct infinite augmentations $\widetilde{\boldsymbol{f}}_{n, m}^s$, it is easy to obtain an upper bound of $\mathcal{L}_M$ as $\mathcal{L}_{PTFA}$, which can be written as:
\begin{equation}
\label{eq_TSFA}
    \lim _{M \rightarrow \infty} \mathcal{L}_M \leq \mathcal{L}_{PTFA} = \frac{1}{N_s} \sum_{n=1}^{N_s}-\log \left( \frac{\vartheta_{n, s, t}^{y_n^s}}{\sum_{c=1}^C\vartheta_{n, s, t}^{c}}\right),
\end{equation}
where $\mathcal{L}_{PTFA}$ is a standard cross-entropy (CE) loss with
\begin{equation}
\label{eq_TSFA_detail}
\begin{aligned}
    \vartheta_{n, s, t}^{c} & = \boldsymbol{w}_{c}^{\top}\boldsymbol{f}_n^s + \left(\boldsymbol{w}_{c}^{\top} - \boldsymbol{w}_{y_{n}^{s}}^{\top}\right)\Delta \mu_t^c \\ & + \frac{\lambda}{2}\left( \boldsymbol{w}_{c}^{\top} - \boldsymbol{w}_{y_{n}^{s}}^{\top}\right)\boldsymbol{\Sigma}_t^c \left( \boldsymbol{w}_{c}^{s} - \boldsymbol{w}_{y_{n}^{s}}\right).
\end{aligned}
\end{equation}
Note that $\boldsymbol{w}_{c}^{\top}\boldsymbol{f}_n^s$ indicates the $c$-th prediction score of $\xi_n^s$. It is efficient to perform classifier adaptation by directly optimizing $\mathcal{L}_{PTFA}$. 

It should be noted that different from TSA~\cite{li2021transferable}, we divide the training process into several stages, and update the pools $\boldsymbol{\mathbb{F}}$, $\boldsymbol{\mathbb{H}}$ and $\boldsymbol{\mathbb{Z}}$ as well as $\boldsymbol{\mu}_s^c$ and $\boldsymbol{\Sigma}_s^c$, $\boldsymbol{\mu}_t^c$ and $\boldsymbol{\Sigma}_t^c$ at the beginning of each stage, while each training stage consists of $\tau$ epochs. In this way, we can ensure a fixed augmentation direction during each stage, which keeps the training stable.

\subsection{Intermediate Domain Feature Extractor Adaptation}
\label{sec:IDA}   
Our PTFA performs classifier adaptation, a key issue of which is the estimation of the augmentation direction that relies heavily on the accuracy of the target predictions. Therefore, we further propose a new intermediate domain feature extractor adaptation (IDFA) strategy for feature extractor adaptation, which can be combined with our PTFA for a better adaptation performance. In particular, we aim at enabling the source features and the target features to approach the mean of the intermediate domain. Specifically, given the extracted source features $\boldsymbol{f}_n^s$, the extracted target features $\boldsymbol{f}_n^t$, and the estimated mean of the intermediate domain $\boldsymbol{\mu}_k^c$, we encourage $\boldsymbol{f}_n^s$ and $\boldsymbol{f}_n^t$ to approach the $\boldsymbol{\mu}_k^{\eta_n}$ of the corresponding category while staying far from the $\boldsymbol{\mu}_k^c$ of other categories. To achieve this, we formulate our IDFA loss as:
\begin{equation}
    \mathcal{L}_{IDFA}^s = \frac{1}{N_s} \sum_{n=1}^{N_s}-\log\left(\frac{\exp \left(\cos\left(\boldsymbol{f}_n^s, \boldsymbol{\mu}_k^{\eta_n^s}\right) / \kappa\right)}{\sum_{c=1}^C \exp \left(\cos\left(\boldsymbol{f}_n^s, \boldsymbol{\mu}_k^c\right) / \kappa\right)}\right),
\end{equation}
and 
\begin{equation}
    \mathcal{L}_{IDFA}^t = \frac{1}{N_t} \sum_{n=1}^{N_t}-\log\left(\frac{\exp \left(\cos\left(\boldsymbol{f}_n^t, \boldsymbol{\mu}_k^{\eta_n^t}\right) / \kappa\right)}{\sum_{c=1}^C \exp \left(\cos\left(\boldsymbol{f}_n^t, \boldsymbol{\mu}_k^c\right) / \kappa\right)}\right),
\end{equation}
where $\kappa$ is the temperature value~\cite{yue2021prototypical, chen2020simple}. Optimizing $\mathcal{L}_{IDFA}^s$ and $\mathcal{L}_{IDFA}^t$ will on one hand promote feature extractor adaptation, preventing the encoder from generating outlier features, and on the other hand encourage the encode to extract more discriminative features. This enhances the adaptation performance of our PTFA method.

\subsection{Overall Objective}
\label{sec:overall_method}   
To sum up, our newly proposed method consists of a classifier adaptation method PTFA and a feature extractor adaptation method IDFA, and the total loss can be written as:
\begin{equation}
\label{eq_total}
    \mathcal{L} = \alpha\mathcal{L}_{PTFA} + \beta\mathcal{L}_{IDFA}^s + \gamma\mathcal{L}_{IDFA}^t,
\end{equation}
where $\alpha$, $\beta$ and $\gamma$ are the trade-off parameters. 

During our implementation, we introduce a warm-up stage at the beginning of the training process to prevent inaccurate intermediate domain estimation. Specifically, during the warm-up stage, we perform the standard cross-entropy (CE) loss to supervise the source predictions
as:
\begin{equation}
\mathcal{L}_{CE} = -\frac{1}{N_s} \sum_{n=1}^{N_s} \ell^{ce}(\boldsymbol{\xi}_{n}, y^s_n).
\end{equation}
We optimize $\mathcal{L}_{CE}$ to train the model in the warm-up stage, and after the warm-up stage, we optimize \cref{eq_total} for the model training.


\section{Experiments}
\label{sec:exp}
\subsection{Implementation Details}
\label{sec:implement}


\begin{table}[t]
\caption{Comparison of classification accuracies (in \%) with the state-of-the-art 3D UDA methods on the PointDA-10 dataset. The best performance is highlighted in bold and the previous best result is underlined. The abbreviations \textit{FEA}, \textit{CA}, and \textit{PS} correspond to feature extractor adaptation, classifier adaptation, and pseudo-labeling, respectively. $^{\ast}$ denotes the utilization of a more sophisticated pseudo-labeling method that requires more parameters during fine-tuning.}
\centering
\scalebox{0.9}{
\begin{tabular}{l c c c | c c c c c c c}
\hline
\textbf{Methods} & FEA & CA & PS & M $\rightarrow$ S & M $\rightarrow$ S$^\ast$ & S $\rightarrow$ M & S $\rightarrow$ S$^\ast$ & S$^\ast$ $\rightarrow$ M &  S$^\ast$ $\rightarrow$ S & Avg.\\
\hline
Oracle & & & & 93.9 & 78.4 & 96.2 & 78.4 & 96.2 & 93.9 & 89.5 \\
w/o DA & & & & 83.3 & 43.8 & 75.5 & 42.5 & 63.8 & 64.2 & 62.2 \\
\hline
ISDA~\cite{NIPS2019_9426} & & $\checkmark$ & & 84.3 & 52.9 & 82.4 & 50.1 & 67.3 & 70.0 & 67.8 \\
TSA~\cite{li2021transferable} & & $\checkmark$ & & 85.1 & 50.3 & \underline{86.9} & 46.4 & \underline{76.9} & 74.7 & 70.0 \\
DANN~\cite{ganin2016domain} & $\checkmark$ & & & 74.8 & 42.1 & 57.5 & 50.9 & 43.7 & 71.6 & 56.8 \\
PointDAN~\cite{qin2019pointdan} & $\checkmark$ & & & 83.9 & 44.8 & 63.3 & 45.7 & 43.6 & 56.4 & 56.3 \\
RS~\cite{sauder2019self} & $\checkmark$ & & & 79.9 & 46.7 & 75.2 & 51.4 & 71.8 & 71.2 & 66.0 \\
DefRec+PCM~\cite{achituve2021self} & $\checkmark$ & & & 81.7 & 51.8 & 78.6 & 54.5 & 73.7 & 71.1 & 68.6  \\
GAST~\cite{zou2021geometry} & $\checkmark$ & & & 83.9 & 56.7 & 76.4 & 55.0 & 73.4 & 72.2 & 69.5 \\
ImplicitPCDA~\cite{shen2022domain} & $\checkmark$ & & & \underline{85.8} & 55.3 & 77.2 & 55.4 & 73.8 & 72.4 & 70.0 \\
DAPS~\cite{wang2023domain} & $\checkmark$ & & & 84.6 & \underline{\textbf{59.2}} & 77.1 & \underline{56.0} & 73.1 & \underline{76.2} & \underline{70.8} \\
\rowcolor[HTML]{EFEFEF} 
Ours & $\checkmark$ & $\checkmark$ & & \textbf{86.6} & 58.5 & \textbf{87.9} & \textbf{59.7} & \textbf{85.1} & \textbf{80.9} & \textbf{76.5} \\
\hline
GAST~\cite{zou2021geometry} & $\checkmark$ & & $\checkmark$ & 84.8 & 59.8 & 80.8 & 56.7 & 81.1 & 74.9 & 73.0 \\
MLSP~\cite{liang2022point} & $\checkmark$ & & $\checkmark$ & 83.7 & 55.4 & 77.1 & 55.6 & 78.2 & 76.1 & 71.0 \\
ImplicitPCDA~\cite{shen2022domain} & $\checkmark$ & & $\checkmark$ & 86.2 & 58.6 & 81.4 & 56.9 & 81.5 & 74.4 & 73.2 \\
GLRV~\cite{fan2022self} & $\checkmark$ & & $\checkmark$ & 85.4 & 60.4 & 78.8 & 57.7 & 77.8 & 76.2 & 72.7 \\
MLSP~\cite{liang2022point} & $\checkmark$ & & $\checkmark$ & 86.2 & 59.1 & \underline{83.5} & 57.6 & 81.2 & 76.4 & 74.0 \\
FD~\cite{cardace2023self} & $\checkmark$ & & $\checkmark$ & 83.9 & \underline{61.1} & 80.3 & \underline{58.9} & \underline{85.5} & 80.9 & 75.1 \\
DAPS~\cite{wang2023domain} & $\checkmark$ & & $\checkmark$ & 86.9 & 59.7 & 78.7 & 55.5 & 82.0 & 80.5 & 73.9 \\
DAS$^{\ast}$~\cite{wang2023domain} & $\checkmark$ & & $\checkmark$ & \underline{87.2} & 60.5 & 82.4 & 58.1 & 84.8 & \underline{82.3} & \underline{75.9} \\ 
\rowcolor[HTML]{EFEFEF} 
Ours & $\checkmark$ & $\checkmark$ & $\checkmark$ & \textbf{88.0} & \textbf{61.7} & \textbf{89.3} & \textbf{60.2} & \textbf{89.0} & \textbf{84.4} & \textbf{78.8} \\
\hline
\end{tabular}
}
\label{table_PointDA}
\end{table}

Following the previous works~\cite{achituve2021self, shen2022domain, zou2021geometry, wang2023domain}, we adopt DGCNN~\cite{wang2019dynamic} as our backbone but simplify the last classifier to a single linear layer, following TSA~\cite{li2021transferable}. In addition, we use a simple max-pooling operation to replace the combination of a max-pooling operation and an average-pooling operation to avoid the over-fitting problem. Our models are trained on a server with four NVIDIA GTX 2080Ti GPUs or a server with ten NVIDIA RTX 3090 GPUs and we only use one GPU per experiment. Our implementation is based on the PyTorch framework. For all experiments, we use the Adam optimizer together with an epoch-wise cosine annealing learning rate scheduler initiated with a learning rate of 0.001 and a weight decay of 0.00005. We train all of our models for 100 epochs setting the numbers of both labeled data and unlabeled data as 8 within each mini-batch. Notably, the 100 training epochs include a 10-epoch warm-up stage. Each intermediate domain construction stage consists of 5 epochs ($\tau=5$). The initial ratio for selecting source data ($\sigma^s$) is set as 1.0, while the initial ratio for selecting target data ($\sigma^t$) is set as 0.0 and the proportional changes $\Delta \sigma^s$ and $\Delta \sigma^t$ are both set as 0.05. The temperature parameter $\kappa$ is set as 2.0, and $\lambda_0$ that controls the semantic variants is set as 0.25, which is adopted from TSA~\cite{li2021transferable}. The trade-off parameters $\alpha$, $\beta$, and $\gamma$ are all set as 1.0, respectively. In addition, we fine-tune our learned model under a self-paced learning (SPL) paradigm where we set the initial threshold for selecting confident predictions as 0.8, and the increasing step as 0.01. The fine-tuning stage consists of 10 training circles and each circle consists of 10 epochs.

\subsection{Experimental Results}
Following previous works~\cite{zou2021geometry, shen2022domain, wang2023domain}, we adopt the widely used 3D unsupervised domain adaptation (UDA) classification benchmark datasets \textbf{PointDA-10 dataset}~\cite{qin2019pointdan} and \textbf{GraspNetPC-10 dataset}~\cite{shen2022domain} to validate the effectiveness of our method.

\begin{table}[t]
\caption{Comparison of classification accuracies (in \%) with the state-of-the-art 3D UDA methods on the GraspNetPC-10 dataset. The best performance is highlighted in bold and the previous best result is underlined. The abbreviations \textit{FEA}, \textit{CA}, and \textit{PS} correspond to feature extractor adaptation, classifier adaptation, and pseudo-labeling, respectively. $^{\ast}$ denotes the utilization of a more sophisticated pseudo-labeling method that requires more parameters during fine-tuning.}
\centering
\scalebox{0.9}{
\begin{tabular}{l c c c | c c c c c}
\hline
\textbf{Methods} & FEA & CA & PS & Syn. $\rightarrow$ Kin. & Syn. $\rightarrow$ RS. & Kin. $\rightarrow$ RS. & RS. $\rightarrow$ Kin. & Avg.\\
\hline
Oracle & & & & 97.2 & 95.6 & 95.6 & 97.2 & 96.4 \\
w/o DA & & & & 61.3 & 54.4 & 53.4 & 68.5 & 59.4 \\
\hline
ISDA~\cite{NIPS2019_9426} & & $\checkmark$ & & 79.5 & 65.0 & 65.3 & 80.8 & 72.7 \\
TSA~\cite{li2021transferable} & & $\checkmark$ & & 77.8 & 60.0 & 57.7 & 85.2 & 70.2 \\
DANN~\cite{ganin2016domain} & $\checkmark$ & & & 78.6 & 70.3 & 46.1 & 67.9 & 65.7 \\
PointDAN~\cite{qin2019pointdan} & $\checkmark$ & & & 77.0 & 72.5 & 65.9 & 82.3 & 74.4 \\
RS~\cite{sauder2019self} & $\checkmark$ & & & 67.3 & 58.6 & 55.7 & 69.6 & 62.8 \\
DefRec+PCM~\cite{achituve2021self} & $\checkmark$ & & & 80.7 & 70.5 & 65.1 & 77.7 & 73.5 \\
GAST~\cite{zou2021geometry} & $\checkmark$ & & & 69.8 & 61.3 & 58.7 & 70.6 & 65.1 \\
ImplicitPCDA~\cite{shen2022domain} & $\checkmark$ & & & 81.2 & 73.1 & 66.4 & 82.6 & 75.8 \\
DAPS~\cite{wang2023domain} & $\checkmark$ & & & \underline{91.6} & \underline{74.2} & \underline{71.9} & \underline{85.0} & \underline{80.7} \\
\rowcolor[HTML]{EFEFEF} 
Ours & $\checkmark$ & $\checkmark$ & & \textbf{94.2} & \textbf{83.5} & \textbf{75.9} & \textbf{96.8} & \textbf{87.6} \\
\hline
GAST~\cite{zou2021geometry} & $\checkmark$ & & $\checkmark$ & 81.3 & 72.3 & 61.3 & 80.1 & 73.8 \\
ImplicitPCDA~\cite{shen2022domain} & $\checkmark$ & & $\checkmark$ & 94.6 & 80.5 & 76.8 & 85.9 & 84.4 \\
DAPS~\cite{wang2023domain} & $\checkmark$ & & $\checkmark$ & 97.0 & 79.6 & 79.1 & 95.7 & 87.8 \\
DAS$^{\ast}$~\cite{wang2023domain} & $\checkmark$ & & $\checkmark$ & \underline{97.2} & \underline{84.4} & \underline{79.9} & \underline{97.0} & \underline{89.6} \\
\rowcolor[HTML]{EFEFEF} 
Ours & $\checkmark$ & $\checkmark$ & $\checkmark$ & \textbf{97.5} & \textbf{84.9} & \textbf{80.1} & \textbf{97.9} & \textbf{90.1} \\
\hline
\end{tabular}
}
\label{table_GraspNet}
\end{table}

We compare our method with recent state-of-the-art (SOTA) UDA methods, including feature extractor adaptation methods (FEA) like DANN~\cite{ganin2016domain} and PointDAN~\cite{qin2019pointdan}, RS~\cite{sauder2019self}, DefRec+PCM~\cite{achituve2021self}, GAST~\cite{zou2021geometry}, ImplicitPCDA~\cite{shen2022domain}, GLRV~\cite{fan2022self}, MLSP~\cite{liang2022point}, FD~\cite{cardace2023self}, DAS~\cite{wang2023domain}, and classifier adaptation methods (CA) like ISDA~\cite{NIPS2019_9426} and TSA~\cite{li2021transferable}. As some methods adopt a pseudo-labeling method (PS) for the model fine-tuning, we report their performance both with and without using PS fine-tuning, corresponding to the lower and upper portions of~\cref{table_PointDA} and~\cref{table_GraspNet}, respectively. It is noteworthy that as DAS employed a more sophisticated PS method, which requires more parameters during training or inference, we also report its performance with the traditional PS method, denoted as DAPS. In addition, we compare with the supervised learning method directly training the model with only labeled source data for reference (denoted as ``w/o DA''), as well as the Oracle method that trains the model by using labeled target data (denoted as ``Oracle'').

The results on the PointDA-10 dataset are reported in~\cref{table_PointDA}. As can be seen, our method surpasses the current SOTA methods by a large margin under all six adaptation scenarios. The average recognition accuracy of our method outperforms the current SOTA method DAS by a notable margin of 2.9\%, while DAS employs a more sophisticated PS method to fine-tune the performance. Our advantage over DAS can be further enlarged to 4.9\% or 5.7\% when comparing ours with DAS using the traditional PS (\ie, DAPS) or without PS, respectively. Moreover, our method demonstrates notable superiority over the two feature augmentation-based methods ISDA~\cite{NIPS2019_9426} and TSA~\cite{li2021transferable}, with improvements of 8.7\% over ISDA and 6.5\% over TSA. This verifies the effectiveness of our progressive classifier and feature extractor adaptation strategy. 

The results on the GraspNetPC-10 dataset are reported in~\cref{table_GraspNet}. Consistently, our method surpasses the w/o DA method by over 30\% and also outperforms the current SOTA method DAS, despite that DAS requires more parameters for a more complicated PS fine-tuning. Our method without PS fine-tuning wins the second-best performer DAPS by a large margin of 6.9\%. Moreover, compared with the two feature augmentation-based methods ISDA and TSA, our method demonstrated overwhelming advantages with almost or more than 15\% improvements, respectively.


Note that our method surpasses the current feature extractor adaptation methods and classifier adaptation methods by a large margin, indicating the great potential of our method in tackling UDA problems.

\subsection{Ablation Study}
\label{sec:ablation}
In this section, we single out the contributions of our module designs on the PointDA-10 dataset. Note that our method includes a progressive target-styled feature augmentation (PTFA) approach for classifier adaptation and an intermediate domain feature extractor adaptation (IDFA) method for feature extractor adaptation, and our PTFA further consists of a basic target-styled feature augmentation (TSFA) and a progressive target-approaching (PTA) strategy. Also note that our PTA strategy is a general training strategy that can also be combined with our IDFA method. The results of the ablation study are reported in~\cref{table_ablation_module}. It can be clearly seen from the table that the basic TSFA will improve the average accuracy of the backbone DGCNN model by 12.1\%, which surpasses TSA~\cite{li2021transferable} as we adopt a more stable training strategy. Applying our IDFA alone will bring a performance improvement of 7.0\% over the w/o DA method. However, \textit{when combining TSFA and IDFA, the performance drops to only 68.6\%}. The main reason is that \textit{the domain gap between the source domain and the target domain is huge} and the above-mentioned two methods both rely heavily on the estimation of the target domain. Directly adapting the model across such a large domain gap will inevitably lead to \textit{incorrect distribution estimation}, harming the adaptation performance of the model.
It can also be inferred from the table that both the combination of our TFSA with our PTA strategy and the combination of our IDFA with our PTA strategy can bring performance improvements of 1.4\% and 4.8\%, respectively. Combining the three components TSFA, IDFA and PTA will bring further improvements of 0.8\% and 2.5\%, respectively, from TSFA+PTA and IDFA+PTA. More importantly, \textit{combining our PTA with TSFA and IDFA brings a performance improvement of nearly 8\% over the combination of TSFA and IDFA.}
The main reason is that we \textit{decompose the huge domain gap into a set of small domain gaps} by introducing a set of intermediate domains, \textit{ensuring an accurate intermediate domain estimation}. These results clearly verify the importance and effectiveness of our PTA strategy. Moreover, it can be seen from the table that only deploying our TSFA and PTA, \ie, our PTFA method, can also outperform the current SOTA method DAPS by a large margin of 4.9\%.

\begin{table}[t]
\caption{Ablation study on the effectiveness of different components in our method, including the basic target-styled feature augmentation (TSFA), our progressive target-approaching (PTA) strategy, and our intermediate domain feature extractor adaptation (IDFA) strategy. Experiments are conducted on the PointDA-10 dataset.}
\centering
\scalebox{0.95}{
\begin{tabular}{c c c | c c c c c c c}
\hline
TSFA & IDFA & PTA & M $\rightarrow$ S & M $\rightarrow$ S$^\ast$  &  S $\rightarrow$ M & S $\rightarrow$ S$^\ast$ & S$^\ast$ $\rightarrow$ M &  S$^\ast$ $\rightarrow$ S & Avg.\\
\hline
& & & 83.3 & 43.8 & 75.5 & 42.5 & 63.8 & 64.2 & 62.2 \\
$\checkmark$ & & & 86.8 & 57.4 & 85.5 & 56.0 & 80.8 & 79.3 & 74.3 \\
& $\checkmark$ & & 85.3 & 52.2 & 80.7 & 55.8 & 68.5 & 72.6 & 69.2 \\
$\checkmark$ & $\checkmark$ & & 84.2 & 54.3 & 81.3 & 55.0 & 65.5 & 71.3 & 68.6 \\
\hline
$\checkmark$ & & $\checkmark$ & 87.5 & 58.2 & 88.4 & 58.3 & 82.1 & 79.4 & 75.7 \\
& $\checkmark$ & $\checkmark$ & 86.6 & 55.7 & 84.7 & 55.7 & 81.9 & 79.1 & 74.0 \\
$\checkmark$ & $\checkmark$ & $\checkmark$ & 86.6 & 58.5 & 87.9 & 59.7 & 85.1 & 80.9 & 76.5 \\
\hline
\end{tabular}
}
\label{table_ablation_module}
\end{table}

\begin{table}[t]
\caption{Ablation study on the effectiveness of different strategies related to our progressive target-approaching (PTA) strategy, including progressively decreasing the selected source samples (decreasing source) and progressively increasing the selected target samples (increasing target). Experiments are conducted on the PointDA-10 dataset. Note that the IDFA module is not included.}
\centering
\scalebox{0.925}{
\begin{tabular}{c c | c c c c c c c}
\hline
decreasing source & increasing target & M $\rightarrow$ S & M $\rightarrow$ S$^\ast$  &  S $\rightarrow$ M & S $\rightarrow$ S$^\ast$ & S$^\ast$ $\rightarrow$ M &  S$^\ast$ $\rightarrow$ S & Avg.\\
\hline
 & & 86.8 & 57.4 & 85.5 & 56.0 & 80.8 & 79.3 & 74.3 \\
 & $\checkmark$ & 87.5 & 58.4 & 85.9 & 56.9 & 79.3 & 77.3 & 74.7 \\
$\checkmark$ & & 87.1 & 56.3 & 89.1 & 58.1 & 82.1 & 79.8 & 75.4 \\
$\checkmark$ & $\checkmark$ & 87.5 & 58.2 & 88.4 & 58.3 & 82.1 & 79.4 & 75.7 \\
\hline
\end{tabular}
}
\label{table_ablation_progressive}
\end{table}

We further delve into the intricate design of our PTA strategy. As a reminder, our approach involves a continual reduction in the number of selected source samples and a simultaneous increase in the number of selected target samples throughout the training process. We thereby scrutinize the effectiveness of each strategy, as reported in~\cref{table_ablation_progressive}. 
As seen, when decreasing the number of the selected source samples while keeping all of the target samples for intermediate domain construction, the model would gradually approach the target domain. In particular, our ``decreasing source'' strategy exhibits great performance under Real-to-Sim scenarios like S$^\ast$ $\rightarrow$ M and S$^\ast$ $\rightarrow$ S. The main reason could be that the samples from the source domain contain noise, which can potentially introduce disruptive information and undermine the model's recognition capabilities. Progressively reducing these samples will help prevent the model from being misled by the noise. 
Moreover, when increasing the number of the selected target samples while keeping all source samples for intermediate domain construction, the adaptation performance on the Sim-to-Real scenarios like M $\rightarrow$ S$^\ast$ and S $\rightarrow$ S$^\ast$ will be improved. The main reason could be that our ``increasing target'' strategy would expand the decision region of the classifier, thus improving the recognition performance of the model on the target domain. The results clearly verified the effectiveness of our PTA strategy.

\begin{figure}[t]
\centering
\includegraphics[width=0.75\linewidth]{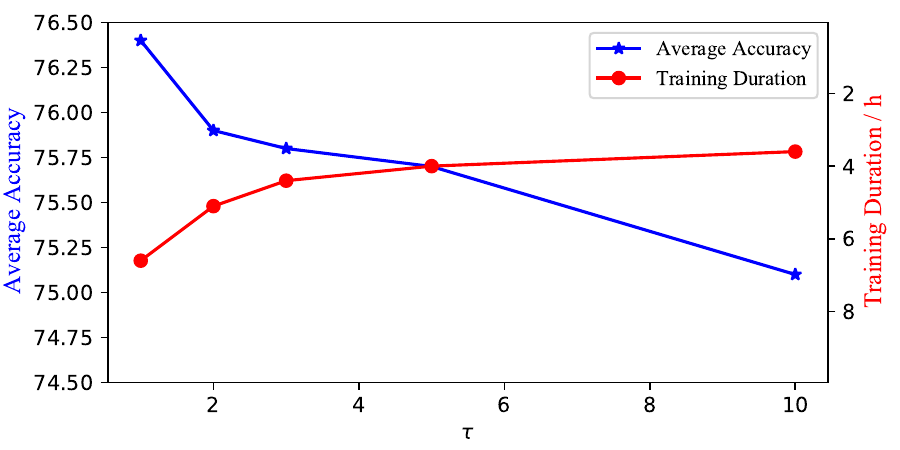}
\caption{Ablation study on the sensitivity of our method to the training epoch at each intermediate domain construction stage, \ie, $\tau$. Experiments are conducted on the PointDA-10 dataset. We report the average recognition accuracy on the target domain and the average training duration.}
\label{fig_stage}
\end{figure}

In addition, we analyze the sensitivity of our method to $\tau$, as depicted in~\cref{fig_stage}. It is evident that a more frequent intermediate domain construction stage results in improved adaptation performance, as it allows for a more precise estimation of augmentation direction. However, the high update frequency comes with increased training costs, as it requires an additional inference stage at the beginning of each intermediate domain construction phase to estimate the augmentation direction. 
Taking into account both training costs and performance, we perform the intermediate construction stage every 5 epochs.






\section{Conclusion}
\label{sec:conclusion}
In this work, we have presented a new progressive classifier and feature extractor adaptation method to tackle unsupervised domain adaptation on point clouds, where we come up with a new progressive target-styled feature augmentation method to tackle the problem from the perspective of classifier adaptation and a new intermediate domain feature extractor adaptation method to tackle the problem from the perspective of feature extractor adaptation. The proposed two methods can mutually benefit each other.
Extensive experiments on the benchmark datasets have validated the effectiveness of our approaches, where our method outperforms the current state-of-the-art methods by a large margin.

\section*{Acknowledgements}
\label{sec:acknowledgements}
This work was supported in part by the Hong Kong Research Grants Council General Research Fund (17203023), in part by The Hong Kong Jockey Club Charities Trust under Grant 2022-0174, in part by the Startup Fund and the Seed Fund for Basic Research for New Staff from The University of Hong Kong, and in part by the funding from UBTECH Robotics.

%
%
\bibliographystyle{splncs04}
\bibliography{main}

\begin{thebibliography}{10}
\providecommand{\url}[1]{\texttt{#1}}
\providecommand{\urlprefix}{URL }
\providecommand{\doi}[1]{https://doi.org/#1}

\bibitem{abnar2021gradual}
Abnar, S., Berg, R.v.d., Ghiasi, G., Dehghani, M., Kalchbrenner, N., Sedghi, H.: Gradual domain adaptation in the wild: When intermediate distributions are absent. arXiv preprint arXiv:2106.06080  (2021)

\bibitem{achituve2021self}
Achituve, I., Maron, H., Chechik, G.: Self-supervised learning for domain adaptation on point clouds. In: Proceedings of the IEEE/CVF Winter Conference on Applications of Computer Vision. pp. 123--133 (2021)

\bibitem{cardace2023self}
Cardace, A., Spezialetti, R., Ramirez, P.Z., Salti, S., Di~Stefano, L.: Self-distillation for unsupervised 3d domain adaptation. In: Proceedings of the IEEE/CVF Winter Conference on Applications of Computer Vision. pp. 4166--4177 (2023)

\bibitem{chen2019progressive}
Chen, C., Xie, W., Huang, W., Rong, Y., Ding, X., Huang, Y., Xu, T., Huang, J.: Progressive feature alignment for unsupervised domain adaptation. In: Proceedings of the IEEE/CVF conference on computer vision and pattern recognition. pp. 627--636 (2019)

\bibitem{chen2020simple}
Chen, T., Kornblith, S., Norouzi, M., Hinton, G.: A simple framework for contrastive learning of visual representations. In: International conference on machine learning. pp. 1597--1607. PMLR (2020)

\bibitem{chen2022exploiting}
Chen, Z., Gu, S., Lu, G., Xu, D.: Exploiting intra-slice and inter-slice redundancy for learning-based lossless volumetric image compression. IEEE Transactions on Image Processing  \textbf{31},  1697--1707 (2022)

\bibitem{chen2021improving}
Chen, Z., Gu, S., Zhu, F., Xu, J., Zhao, R.: Improving facial attribute recognition by group and graph learning. In: 2021 IEEE International Conference on Multimedia and Expo (ICME). pp.~1--6. IEEE (2021)

\bibitem{chen2022lsvc}
Chen, Z., Lu, G., Hu, Z., Liu, S., Jiang, W., Xu, D.: Lsvc: A learning-based stereo video compression framework. In: Proceedings of the IEEE/CVF Conference on Computer Vision and Pattern Recognition. pp. 6073--6082 (2022)

\bibitem{chen2023neural}
Chen, Z., Relic, L., Azevedo, R., Zhang, Y., Gross, M., Xu, D., Zhou, L., Schroers, C.: Neural video compression with spatio-temporal cross-covariance transformers. In: Proceedings of the 31st ACM International Conference on Multimedia. pp. 8543--8551 (2023)

\bibitem{chu2023towards}
Chu, M., Zheng, Z., Ji, W., Chua, T.S.: Towards natural language-guided drones: Geotext-1652 benchmark with spatially relation matching. arXiv preprint arXiv:2311.12751  (2023)

\bibitem{das2018embodied}
Das, A., Datta, S., Gkioxari, G., Lee, S., Parikh, D., Batra, D.: Embodied question answering. In: Proceedings of the IEEE conference on computer vision and pattern recognition. pp. 1--10 (2018)

\bibitem{fan2022self}
Fan, H., Chang, X., Zhang, W., Cheng, Y., Sun, Y., Kankanhalli, M.: Self-supervised global-local structure modeling for point cloud domain adaptation with reliable voted pseudo labels. In: Proceedings of the IEEE/CVF Conference on Computer Vision and Pattern Recognition. pp. 6377--6386 (2022)

\bibitem{fang2020graspnet}
Fang, H.S., Wang, C., Gou, M., Lu, C.: Graspnet-1billion: A large-scale benchmark for general object grasping. In: Proceedings of the IEEE/CVF conference on computer vision and pattern recognition. pp. 11444--11453 (2020)

\bibitem{ganin2016domain}
Ganin, Y., Ustinova, E., Ajakan, H., Germain, P., Larochelle, H., Laviolette, F., Marchand, M., Lempitsky, V.: Domain-adversarial training of neural networks. The journal of machine learning research  \textbf{17}(1),  2096--2030 (2016)

\bibitem{ghifary2016deep}
Ghifary, M., Kleijn, W.B., Zhang, M., Balduzzi, D., Li, W.: Deep reconstruction-classification networks for unsupervised domain adaptation. In: Computer Vision--ECCV 2016: 14th European Conference, Amsterdam, The Netherlands, October 11--14, 2016, Proceedings, Part IV 14. pp. 597--613. Springer (2016)

\bibitem{hsu2020progressive}
Hsu, H.K., Yao, C.H., Tsai, Y.H., Hung, W.C., Tseng, H.Y., Singh, M., Yang, M.H.: Progressive domain adaptation for object detection. In: Proceedings of the IEEE/CVF winter conference on applications of computer vision. pp. 749--757 (2020)

\bibitem{kumar2020understanding}
Kumar, A., Ma, T., Liang, P.: Understanding self-training for gradual domain adaptation. In: International conference on machine learning. pp. 5468--5479. PMLR (2020)

\bibitem{li2021transferable}
Li, S., Xie, M., Gong, K., Liu, C.H., Wang, Y., Li, W.: Transferable semantic augmentation for domain adaptation. In: Proceedings of the IEEE/CVF conference on computer vision and pattern recognition. pp. 11516--11525 (2021)

\bibitem{li2013learning}
Li, W., Duan, L., Xu, D., Tsang, I.W.: Learning with augmented features for supervised and semi-supervised heterogeneous domain adaptation. IEEE Transactions on Pattern analysis and machine intelligence  \textbf{36}(6),  1134--1148 (2013)

\bibitem{liang2022point}
Liang, H., Fan, H., Fan, Z., Wang, Y., Chen, T., Cheng, Y., Wang, Z.: Point cloud domain adaptation via masked local 3d structure prediction. In: European Conference on Computer Vision. pp. 156--172. Springer (2022)

\bibitem{lifshitz2021sample}
Lifshitz, O., Wolf, L.: Sample selection for universal domain adaptation. In: Proceedings of the AAAI Conference on Artificial Intelligence. vol.~35, pp. 8592--8600 (2021)

\bibitem{liu2024lta}
Liu, J., Li, J., Wang, K., Guo, H., Yang, J., Peng, J., Xu, K., Liu, X., Guo, J.: Lta-pcs: Learnable task-agnostic point cloud sampling. In: Proceedings of the IEEE/CVF Conference on Computer Vision and Pattern Recognition. pp. 28035--28045 (2024)

\bibitem{qi2017pointnet}
Qi, C.R., Su, H., Mo, K., Guibas, L.J.: Pointnet: Deep learning on point sets for 3d classification and segmentation. In: Proceedings of the IEEE conference on computer vision and pattern recognition. pp. 652--660 (2017)

\bibitem{qi2017pointnet++}
Qi, C.R., Yi, L., Su, H., Guibas, L.J.: Pointnet++: Deep hierarchical feature learning on point sets in a metric space. Advances in neural information processing systems  \textbf{30} (2017)

\bibitem{qin2019pointdan}
Qin, C., You, H., Wang, L., Kuo, C.C.J., Fu, Y.: Pointdan: A multi-scale 3d domain adaption network for point cloud representation. Advances in Neural Information Processing Systems  \textbf{32} (2019)

\bibitem{rao2020global}
Rao, Y., Lu, J., Zhou, J.: Global-local bidirectional reasoning for unsupervised representation learning of 3d point clouds. In: Proceedings of the IEEE/CVF Conference on Computer Vision and Pattern Recognition. pp. 5376--5385 (2020)

\bibitem{rao2022pointglr}
Rao, Y., Lu, J., Zhou, J.: Pointglr: Unsupervised structural representation learning of 3d point clouds. IEEE Transactions on Pattern Analysis and Machine Intelligence  (2022)

\bibitem{rukhovich2023tr3d}
Rukhovich, D., Vorontsova, A., Konushin, A.: Tr3d: Towards real-time indoor 3d object detection. arXiv preprint arXiv:2302.02858  (2023)

\bibitem{sauder2019self}
Sauder, J., Sievers, B.: Self-supervised deep learning on point clouds by reconstructing space. Advances in Neural Information Processing Systems  \textbf{32} (2019)

\bibitem{shen2022domain}
Shen, Y., Yang, Y., Yan, M., Wang, H., Zheng, Y., Guibas, L.J.: Domain adaptation on point clouds via geometry-aware implicits. In: Proceedings of the IEEE/CVF Conference on Computer Vision and Pattern Recognition. pp. 7223--7232 (2022)

\bibitem{shi2020pv}
Shi, S., Guo, C., Jiang, L., Wang, Z., Shi, J., Wang, X., Li, H.: Pv-rcnn: Point-voxel feature set abstraction for 3d object detection. In: Proceedings of the IEEE/CVF Conference on Computer Vision and Pattern Recognition. pp. 10529--10538 (2020)

\bibitem{shi2019pointrcnn}
Shi, S., Wang, X., Li, H.: Pointrcnn: 3d object proposal generation and detection from point cloud. In: Proceedings of the IEEE/CVF conference on computer vision and pattern recognition. pp. 770--779 (2019)

\bibitem{tang2021vicinal}
Tang, H., Jia, K.: Vicinal and categorical domain adaptation. Pattern Recognition  \textbf{115},  107907 (2021)

\bibitem{wang2021cross}
Wang, F., Li, W., Xu, D.: Cross-dataset point cloud recognition using deep-shallow domain adaptation network. IEEE Transactions on Image Processing  \textbf{30},  7364--7377 (2021)

\bibitem{wang2019dynamic}
Wang, Y., Sun, Y., Liu, Z., Sarma, S.E., Bronstein, M.M., Solomon, J.M.: Dynamic graph cnn for learning on point clouds. ACM Transactions on Graphics (tog)  \textbf{38}(5),  1--12 (2019)

\bibitem{wang2021regularizing}
Wang, Y., Huang, G., Song, S., Pan, X., Xia, Y., Wu, C.: Regularizing deep networks with semantic data augmentation. IEEE Transactions on Pattern Analysis and Machine Intelligence  (2021). \doi{10.1109/TPAMI.2021.3052951}

\bibitem{NIPS2019_9426}
Wang, Y., Pan, X., Song, S., Zhang, H., Huang, G., Wu, C.: Implicit semantic data augmentation for deep networks. In: Advances in Neural Information Processing Systems (NeurIPS). pp. 12635--12644 (2019)

\bibitem{wang2024pointramba}
Wang, Z., Chen, Z., Wu, Y., Zhao, Z., Zhou, L., Xu, D.: Pointramba: A hybrid transformer-mamba framework for point cloud analysis. arXiv preprint arXiv:2405.15463  (2024)

\bibitem{wang2023domain}
Wang, Z., Li, W., Xu, D.: Domain adaptive sampling for cross-domain point cloud recognition. IEEE Transactions on Circuits and Systems for Video Technology  (2023)

\bibitem{wang2023conflict}
Wang, Z., Zhao, Z., Xing, X., Xu, D., Kong, X., Zhou, L.: Conflict-based cross-view consistency for semi-supervised semantic segmentation. In: Proceedings of the IEEE/CVF conference on computer vision and pattern recognition. pp. 19585--19595 (2023)

\bibitem{wu2023virtual}
Wu, H., Wen, C., Shi, S., Li, X., Wang, C.: Virtual sparse convolution for multimodal 3d object detection. In: Proceedings of the IEEE/CVF Conference on Computer Vision and Pattern Recognition. pp. 21653--21662 (2023)

\bibitem{wu2021mgh}
Wu, Y., Wu, X., Li, X., Tian, J.: Mgh: Metadata guided hypergraph modeling for unsupervised person re-identification. In: Proceedings of the 29th ACM International Conference on Multimedia. pp. 1571--1580 (2021)

\bibitem{yue2021prototypical}
Yue, X., Zheng, Z., Zhang, S., Gao, Y., Darrell, T., Keutzer, K., Vincentelli, A.S.: Prototypical cross-domain self-supervised learning for few-shot unsupervised domain adaptation. In: Proceedings of the IEEE/CVF Conference on Computer Vision and Pattern Recognition. pp. 13834--13844 (2021)

\bibitem{zhang2018collaborative}
Zhang, W., Ouyang, W., Li, W., Xu, D.: Collaborative and adversarial network for unsupervised domain adaptation. In: Proceedings of the IEEE conference on computer vision and pattern recognition. pp. 3801--3809 (2018)

\bibitem{zhang2021gradual}
Zhang, Y., Deng, B., Jia, K., Zhang, L.: Gradual domain adaptation via self-training of auxiliary models. arXiv preprint arXiv:2106.09890  (2021)

\bibitem{zhao2021mgsvf}
Zhao, H., Fu, Y., Kang, M., Tian, Q., Wu, F., Li, X.: Mgsvf: Multi-grained slow vs. fast framework for few-shot class-incremental learning. IEEE Transactions on Pattern Analysis and Machine Intelligence  (2021)

\bibitem{zhao2021video}
Zhao, H., Qin, X., Su, S., Fu, Y., Lin, Z., Li, X.: When video classification meets incremental classes. In: Proceedings of the 29th ACM International Conference on Multimedia. pp. 880--889 (2021)

\bibitem{zhao2021memory}
Zhao, H., Wang, H., Fu, Y., Wu, F., Li, X.: Memory-efficient class-incremental learning for image classification. IEEE Transactions on Neural Networks and Learning Systems  \textbf{33}(10),  5966--5977 (2021)

\bibitem{zhao2022rbc}
Zhao, H., Yang, F., Fu, X., Li, X.: Rbc: Rectifying the biased context in continual semantic segmentation. In: European Conference on Computer Vision. pp. 55--72. Springer (2022)

\bibitem{zhao2023instance}
Zhao, Z., Long, S., Pi, J., Wang, J., Zhou, L.: Instance-specific and model-adaptive supervision for semi-supervised semantic segmentation. In: Proceedings of the IEEE/CVF conference on computer vision and pattern recognition. pp. 23705--23714 (2023)

\bibitem{zhao2023alternate}
Zhao, Z., Wang, Z., Wang, L., Yuan, Y., Zhou, L.: Alternate diverse teaching for semi-supervised medical image segmentation. arXiv preprint arXiv:2311.17325  (2023)

\bibitem{zhao2023augmentation}
Zhao, Z., Yang, L., Long, S., Pi, J., Zhou, L., Wang, J.: Augmentation matters: A simple-yet-effective approach to semi-supervised semantic segmentation. In: Proceedings of the IEEE/CVF conference on computer vision and pattern recognition. pp. 11350--11359 (2023)

\bibitem{zhao2022dc}
Zhao, Z., Zhou, L., Duan, Y., Wang, L., Qi, L., Shi, Y.: Dc-ssl: Addressing mismatched class distribution in semi-supervised learning. In: Proceedings of the IEEE/CVF conference on computer vision and pattern recognition. pp. 9757--9765 (2022)

\bibitem{zhao2022lassl}
Zhao, Z., Zhou, L., Wang, L., Shi, Y., Gao, Y.: Lassl: Label-guided self-training for semi-supervised learning. In: Proceedings of the AAAI conference on artificial intelligence. vol.~36, pp. 9208--9216 (2022)

\bibitem{zou2021geometry}
Zou, L., Tang, H., Chen, K., Jia, K.: Geometry-aware self-training for unsupervised domain adaptation on object point clouds. In: Proceedings of the IEEE/CVF International Conference on Computer Vision. pp. 6403--6412 (2021)

\end{thebibliography}
\end{document}